\documentclass[runningheads]{llncs}
\usepackage[T1]{fontenc}
\usepackage{caption}
\usepackage{graphicx,verbatim}
\usepackage{amsmath}  
\usepackage{amssymb}  
\usepackage{booktabs}
\usepackage{multirow}
\usepackage{color}
\usepackage{xcolor}
\usepackage{marvosym}
\usepackage[colorlinks=true, 
            linkcolor=blue,   
            citecolor=blue,   
            urlcolor=blue     
            ]{hyperref}

\usepackage{caption}

\begin{document}
\title{Echo4DIR: 4D Implicit Heart Reconstruction from 2D Echocardiography Videos}
%
\author{Yanan Liu\inst{1,2} \and
Qinya Li\inst{1} \and
Hao Zhang\inst{2} \and
Kangjian He\inst{2} \and
Xuan Yang \inst{1} \and
Hao Li \inst{1} \and
Dan Xu \inst{2}\textsuperscript{\Letter} \and
Lei Li \inst{1}\textsuperscript{\Letter}}

\authorrunning{F. Author et al.}
%
\institute{Department of Biomedical Engineering, National University of Singapore, Singapore \and
School of Information Science and Engineering, Yunnan University, Kunming, China}


  
\maketitle              
\begin{abstract}
Reconstructing 4D (3D+t) cardiac geometry from sparse 2D echocardiography is highly desirable yet fundamentally challenged by  geometric ambiguity and temporal discontinuity. To tackle these issues, we propose Echo4DIR, a novel test-time 4D implicit reconstruction framework.
Specifically, we learn robust 3D shape priors from statistical shape models (SSMs) via a cardiac conditional SDF, constructing an  Epipolar Mask Encoder  module with  epipolar cross attention to effectively fuse  multi-view features.
To bridge the synthetic-to-real domain gap, we introduce a self-supervised SDF-tailored differentiable rendering strategy for patient-specific 3D shape adaptation using uncalibrated clinical masks without requiring 3D ground truth. 
Crucially, the inherent continuity of implicit representation overcomes sparse observations, enabling anatomically reliable geometry at arbitrary resolutions.
Furthermore, to empower our framework with physically continuous 4D extension, we introduce a Radial SDF Alignment strategy that strictly locks shape evolution to the predicted velocity field, fundamentally eliminating mesh drift.
Extensive experiments on synthetic benchmarks and real clinical datasets demonstrate that Echo4DIR achieves state-of-the-art 4D cardiac mesh reconstruction, notably yielding an impressive clinical overlap of up to 98.35\% Dice and 96.75\% IoU. The code will be available in \url{https://github.com/ryannus2025-ai/Echo4DIR}.



\end{abstract}
\section{Introduction}

Patient-specific 3D+t cardiac anatomical twins~\cite{cai,UltraTwin}, which enable the automated assessment of cardiac morphology~\cite{cm}, motion~\cite{yxh}, and function~\cite{cf}, hold immense promise for the early diagnosis and surgical navigation of cardiovascular diseases. While cardiac Computed Tomography (CT)~\cite{CCT}, and Magnetic Resonance Imaging (MRI)~\cite{yxh}, provide high-quality static anatomical details, their high cost or potential radiation exposure restrict their application in frequent examinations.  In contrast, 2D Ultrasound (US)~\cite{MIA23}, despite being ubiquitous and real-time, inherently lacks 3D spatial context and relies heavily on clinician experience, introducing significant diagnostic uncertainty. Furthermore, 3D  US~\cite{3DUS} is often compromised by motion artifacts.  Therefore, reconstructing 3D+t cardiac structures from multi-view 2D US emerges as a promising approach to achieve low-cost, high-precision, and personalized cardiac anatomical twins.

Recently, several impressed approaches have emerged.  Stojanovsk et al. ~\cite{Pix}   reconstructed 3D cardiac voxels from multi-view synthetic 2D US, which ignores the synthetic-real  domain shift.  Laumer et al.~\cite{MIA23} proposed a weakly supervised single-view reconstruction framework,  training an image-mesh autoencoder~\cite{CycleGAN} using cycle-consistency loss.  Subsequently, Li et al.~\cite{UltraTwin} introduced a generative framework based on diffusion models~\cite{DiT}, learning the conditional distribution of 3D cardiac anatomy in a latent space with the CT supervision. Yet, the probabilistic nature of such models may lead to anatomical hallucinations, hindering the patient-specific reconstruction. A recent approach~\cite{MIA25} integrated Graph Convolutional Networks (GCNs)~\cite{STGCN} for 3D+t left ventricular (LV) mesh reconstruction, trying to capture the diversity of  clinical cases via explicit differentiable rendering.  Despite these advancements, reconstructing 3D+t cardiac geometry from sparse 2D US views remains the challenges: \textbf{1)} \textit{Geometric Ambiguity.} The inherent sparsity of 2D echocardiography views induces significant information loss, particularly in unseen areas, making the model difficult to determine the underlying 3D shape. 2) \textit{Temporal Discontinuity.} Directly extending static reconstruction to 3D+t sequences fails to establish dense anatomical correspondences across frames. This lack of explicit kinematic constraints leads to temporal artifacts such as mesh flickering and non-physical deformations.

To tackle these challenges, we propose \textbf{Echo4DIR}, a pioneering  \textbf{Echo}-cardiography based \textbf{4D} \textbf{I}mplicit cardiac \textbf{R}econstruction framework via conditional Signed Distance Functions (SDFs)~\cite{SDF}. Specifically, Echo4DIR integrates three key designs: \textbf{1) }We pioneer the integration of conditional SDFs  into echo-based 3D+t cardiac shape reconstruction to optimize geometric ambiguity.  Utilizing multi-view 2D mask features fused by the Epipolar Cross Attention (ECA)~\cite{ECA} as the condition, we learn a continuous 3D cardiac SDF trained by SSM priors.  \textbf{2) }We introduce a Differentiable Rendering (DR) strategy tailored for SDFs to bridge the domain gap between synthetic SSMs and real clinical data, which  fine-tunes the implicit surface using real echo masks and learnable transducer parameters, yielding high-fidelity, patient-specific cardiac geometry. \textbf{3) }We introduce a Radial SDF Alignment (RSA) strategy  driven by the learned velocity field to harmonize spatial shape with temporal motion correlations, which resolves the topological inconsistency (mesh drift and volume collapse) inherent in SDF-derived meshes,  achieving effective 3D+t dynamic reconstruction.
Essentially, our proposed Echo4DIR operates as a Test-Time Optimization (TTO) framework enabling patient-specific shape adaptation. By exploiting the continuity of SDFs, it enables the plausible completion of unobserved geometry at arbitrary resolutions and robustly extends SDFs to the dynamic 3D+t domain. 

\section{Methodology}
\begin{figure}[t]
\centering
\includegraphics[width=\textwidth]{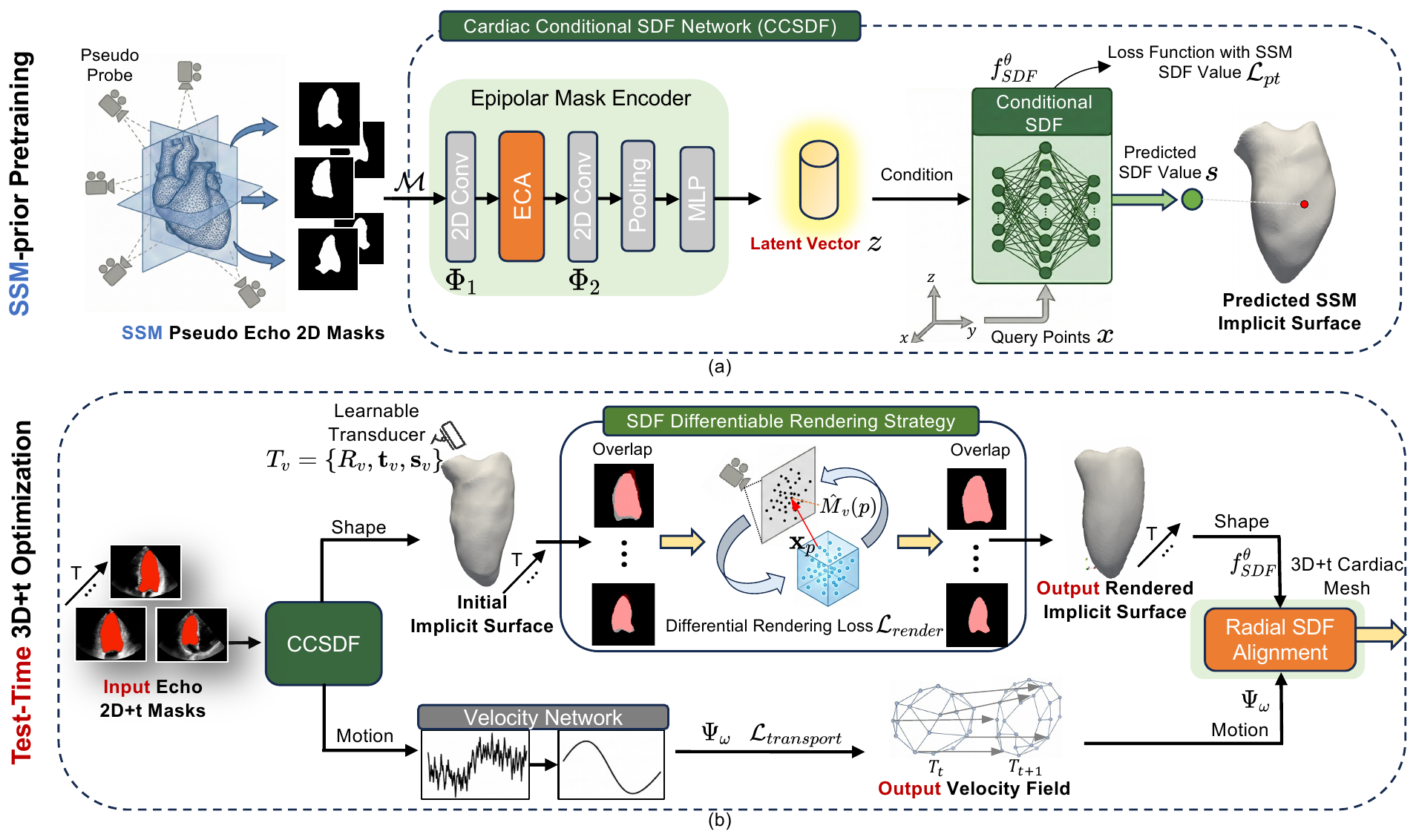}
\caption{Overall Architecture of our proposed  Echo4DIR framework.} \label{fig:OA}
\end{figure}

\textbf{Overall Architecture.} As shown in Fig. \ref{fig:OA}, we propose Echo4DIR, a pioneering 3D+t cardiac reconstruction framework via conditional SDFs from sparse-view 2D echo masks, consisting of: \textbf{1)} SSM-pretrained Cardiac Conditional SDF network (CCSDF) to learn cardiac shape priors. As shown in Fig.~\ref{fig:OA} (a),  we simulate the US probe imaging on SSM meshes to generate  pseudo masks from different views, which are encoded by the Epipolar Mask  Encoder (EME) module integrated with a Epipolar Cross Attention (ECA) into the latent vector as the condition of SDFs.   \textbf{2)} Test-time 3D+t Reconstruction Optimization. As shown in Fig.~\ref{fig:OA} (b),  for shape optimization, clinical 2D echo masks are fed into CCSDF with learnable transducer to initialize the implicit surface.  We introduce a SDF-tailored differentiable rendering strategy to enable self-supervised shape adaptation.  For motion optimization, we introduce the Velocity Network to predict an implicit velocity field for any SDF point over time, integrating with a Radial SDF Alignment strategy to lock shape evolution with the velocity field, further achieving the robust 3D+t reconstruction. Below, we describe the details.

\subsection{3D Cardiac Shape Representation via Conditional Signed Distance Function}

To overcome the discretization artifacts in voxel and  mesh based representations, we introduce the deep-learning conditional Signed Distance Function (SDF)~\cite{SDF}.  Firstly, we define the SDF as a continuous mapping $SDF(\cdot):\mathbb{R}^3 \to \mathbb{R}$. For any spatial query point $\mathbf{x} \in \mathbb{R}^3$, the value $s = SDF(\mathbf{x})\in\mathbb{R}$ represents the Euclidean distance to the nearest surface boundary $\mathcal{S}$, where the sign indicates the interior  or exterior  regions.  The cardiac surface is implicitly represented by the zero-level set $\mathcal{S} = \{ \mathbf{x}\in\mathbb{R}^3\mid s = SDF(\mathbf{x})= 0 \}$.  Next, we encode the individual target shape into a latent vector $\mathbf{z}$ and  parameterize $SDF(\cdot)$ using a Multi-Layer Perceptron (MLP)~\cite{MLP}. The conditional SDF $f^{\theta}_{SDF}$ can be formulated as:
\begin{equation}
    \label{eq1}
f^{\theta}_{SDF}(\mathbf{z}, \mathbf{x})\approx {SDF}(\mathbf{x}), \forall \mathbf{x} \in \Omega,
\end{equation}
where $\theta$ denotes the learnable parameters of the MLP, $\mathbf{z} \in \mathbb{R}^L$ is the latent vector encapsulating the patient-specific geometry and $L$ is the dimension of the latent space. $\Omega \subset \mathbb{R}^3$ represents the set of spatial query points. 

\textbf{Epipolar Mask  Encoder.} We construct an Epipolar Mask Encoder to generate a personalized latent  vector from $v$-view  echo masks. First, a shared-weight 2D convolutional backbone~\cite{ResNet} $\Phi_1$ is used to extract the feature map $F_v=\Phi_v(M_v)\in \mathbb{R}^{C \times H' \times W'}$ from the  mask $M_v$, where $C$, $H'$ and $W'$ are the number of channels, height and width. To capture the cross-view correlations~\cite{ECA}, we introduce an Epipolar Cross Attention (ECA) module, which projects the anchor feature $F_0$ into the Query $Q \in \mathbb{R}^{C \times H' \times W'}$ and projects the auxiliary feature $F_i$ into the Key and Value $K,V\in \mathbb{R}^{C \times H' \times W'}$. Then we define an epipolar vertical window $\mathcal{W}$ to address the probe bias, namely,  each pixel $p$ on the anchor view only needs to search within the corresponding window of the auxiliary view as:
\begin{equation}
\begin{aligned}
&\hat{F}(p) = F_0(p) + \sum_{j \in \mathcal{W}(p)} \text{Softmax}\left(\frac{Q(p) \cdot K(j)}{\tau}\right) V(j),\\
&\mathbf{z} = \text{MLP}\left( \text{AvgPool}\left( \Phi_2(\hat{F}) \right) \right),
\label{eq:latent_z}
\end{aligned}
\end{equation}
where $\tau$ is a temperature  factor.  $\hat{F}$ encodes the  multi-view context. $\Phi_2$ denotes the convolution. $\text{AvgPool}(\cdot)$ and  $\text{MLP}(\cdot)$ denote average pooling and MLP.


\subsection{Self-supervised SDF-tailored Differentiable Rendering}

During test-time optimization, the model encounters clinical 2D echo masks that exhibit inherent domain gaps from the SSM priors. 
To enable patient-specific shape adaptation, we introduce a SDF-tailored differentiable rendering strategy.
To address probe misalignment, 
we parameterize the transducer for each view $v$ via a learnable spatial transformation $T_v = \{R_v, \mathbf{t}_v, \mathbf{s}_v\}$, denoting \textit{rotation}, \textit{translation}, and \textit{scaling}, respectively. This allows us to explicitly map any sampled 2D pixel $p = (x, y)$ on the image plane ($\bar{p} = [x, y, 0]^\top$) to its corresponding 3D spatial coordinate $\mathbf{x}_p = \mathbf{s}_v R_v \bar{p} + \mathbf{t}_v.$

After querying the 3D SDF value at $\mathbf{x}_p$, the continuous distance should be converted into a 2D mask probability to compute the self-supervised loss. Traditional methods for surface extraction (\textit{e.g.}, the step function) is non-differentiable and severs the gradient flow~\cite{MIA25}. To resolve this, we formulate a differentiable rendering strategy that maps the predicted SDF value $f_{SDF}^\theta(\mathbf{z}, \mathbf{x}_p)$ to a soft probability $\hat{M}_v(p)$ using a scaled Sigmoid function with a temperature parameter $\alpha > 0$ as:
\begin{equation}
\hat{M}_v(p) = \frac{1}{1 + \exp(\alpha \cdot f_{SDF}^\theta(\mathbf{z}, \mathbf{x}_p))}.
\label{eq:rendering}
\end{equation}

\subsection{Cardiac Spatial-Temporal Reconstruction via Neural Velocity Modeling}

We define a continuous neural velocity field $\Psi_\omega$ parameterized by a MLP. Given a 3D spatial point $\mathbf{x}_t \in \mathbb{R}^3$ and a normalized timestamp $t \in [0, 1]$, $\Psi_\omega$ predicts the  velocity vector $\mathbf{v}_t\in \mathbb{R}^3$. Here, a sinusoidal positional encoding $\gamma(\cdot)$~\cite{NerF} is applied to preserve high-frequency motion details:
\begin{equation}
\mathbf{v}_t = \Psi_\omega(\gamma(\mathbf{x}), \gamma(t)).
\end{equation}
Furthermore, we employ a bidirectional LSTM~\cite{BiLSTM} $\Phi_{BiL}$ to refine the temporal motion correlations of the latent vectors $[\tilde{\mathbf{z}}_1, \dots, \tilde{\mathbf{z}}_T] = \Phi_{BiL}([\mathbf{z}_1,  \dots, \mathbf{z}_T];\theta_{BiL})$.
Therefore, we obtained the C-SDF $f_{SDF}^\theta$ for 3D shape and the velocity field $\Psi_\omega$ for cardiac motion. To overcome the inherent limitation of the SDF (topologically inconsistent meshes generated by frame-by-frame Marching Cubes~\cite{Marching}),  we extract only a single initial 3D mesh $\mathbf{M}_0 = (\mathcal{X}_0, \mathcal{F})$
at $t=0$ from $f_{SDF}^\theta$, where $\mathcal{X}_0$ and $\mathcal{F}$ denote the initial vertex set and face topology. For subsequent frames, we propagate  each individual vertex $\mathbf{x}_{t-1} \in \mathcal{X}_{t-1}$  using the $\Psi_\omega$ as $\mathbf{\hat{x}}_t = \mathbf{x}_{t-1} + \mathbf{v}_{t-1}$. To correct shape deviations accumulated by $\Psi_\omega$, we introduce an explicit Radial SDF Alignment strategy leveraging the shape prior from each frame's SDF:
\begin{equation}
    \mathbf{x}_t = \mathbf{\hat{x}}_t - f_{SDF}^\theta(\tilde{\mathbf{z}}_t, \mathbf{\hat{x}}_t) \cdot \frac{\nabla f_{SDF}^\theta(\tilde{\mathbf{z}}_t, \mathbf{\hat{x}}_t)}{\|\nabla f_{SDF}^\theta(\tilde{\mathbf{z}}_t, \mathbf{\hat{x}}_t)\|_2}
    \label{eq:sdf_alignment}
\end{equation}
where $\mathbf{\hat{x}}_t$ and $\mathbf{x}_t$ denote the drifted vertices  and the  aligned vertices at frame $t$. The scalar $f_{SDF}^\theta(\tilde{\mathbf{z}}_t, \mathbf{\hat{x}}_t)$ evaluates the signed distance from the drifted vertex to the true surface, while  $\frac{\nabla f_{SDF}^\theta}{\|\nabla f_{SDF}^\theta\|_2}$ represents the unit surface normal. By moving the drifted vertex $\mathbf{\hat{x}}_t$ along the inverse normal direction by the SDF distance, this operation explicitly pushes it onto the zero-level set ($f_{SDF}^\theta=0$).

\subsection{Model Optimization and Loss Function}
We optimize our proposed Echo4DIR using two strategies: 1) SSM priors Supervised  pretraining and 2) Test-time self-supervised learning.

\textbf{SSM Priors Supervised Pretraining.} We pretrain the CCSDF  using SSM data~\cite{MIA23,MIA25}, simulating US imaging to synthesize pseudo-echo masks by rotating the  meshes to standard apical views. 
The CCSDF predicts the implicit surface at sampled 3D query points $\mathbf{x} \in \Omega$. The network is optimized end-to-end using an $L_1$ reconstruction loss and an $L_2$ regularization loss to prevent overfitting as:
\begin{equation}
\mathcal{L}_{pt} = \frac{1}{|\Omega|} \sum_{\mathbf{x} \in \Omega} \left| f^{\theta}_{SDF}(\mathbf{z}, \mathbf{x}) - s_{gt}(\mathbf{x}) \right|_1 + \lambda_1 |\mathbf{z}|_2^2,
\label{eq:pretrain_loss}
\end{equation}
where $s_{gt}$ is the ground-truth SDF from SSM mesh and $\lambda_1$ is the weight scalar. 

\textbf{Test-time self-supervised learning. } 1) \textit{Shape} self-supervised learning. The cardiac shape is optimized by a hybrid loss $\mathcal{L}_{render}$, which combines pixel-wise Binary Cross-Entropy (BCE) $\mathcal{L}_{BCE}$ and global Dice  Loss $\mathcal{L}_{Dice}$ between the rendered probability $\hat{M}_v$ and the clinical mask $M_v$:
\begin{equation}
\mathcal{L}_{render} = \frac{1}{V} \sum_{v=1}^V\Big(  \frac{1}{|\mathcal{P}|} \sum_{p \in \mathcal{P}} \mathcal{L}_{BCE}(\hat{M}_v(p), M_v(p)) + \mathcal{L}_{Dice}(\hat{M}_v, M_v) \Big).\label{eq:hybrid_loss}\end{equation}

2) \textit{Motion} self-supervised learning. We supervise the velocity network $\Psi_\omega$  through the temporal evolving implicit surfaces.  Based on Lagrangian kinematics~\cite{La1,La2}, a physical point $\mathbf{x}_t$ advected by $\mathbf{v}_t$ should lie on the zero-level set of the next frame's surface. To ensure robust  motion tracking, we introduce a bidirectional transport  loss~\cite{La1} as:
\begin{equation}
\begin{aligned}
\mathcal{L}_{transport} = &\mathbb{E}_{\mathbf{x}_t} \Big[ w(\mathbf{x}_t) \cdot | f_{SDF}^\theta(\tilde{\mathbf{z}}_{t+1}, \mathbf{x}_t + \mathbf{v}_t) |_2^2 \Big] + \\
&\mathbb{E}_{\mathbf{x}_{t+1}} \Big[ w(\mathbf{x}_{t+1}) \cdot | f_{SDF}^\theta(\tilde{\mathbf{z}}_t, \mathbf{x}_{t+1} - \mathbf{v}_{t+1}) |_2^2 \Big]+ \mathcal{L}_{smooth},
\end{aligned}
\end{equation}
where the weight $w(\mathbf{x}) = \exp(- |f_{SDF}^\theta(\tilde{\mathbf{z}}, \mathbf{x})|)$ ensures the optimization focuses on the anatomical boundaries. $\mathbb{E}_{\mathbf{x}_{t}}$ denotes the  expectation over the sampled points $\mathbf{x}_{t}$.  The smoothness loss  $\mathcal{L}_{smooth}=\mathbb{E}_{\mathbf{x} \sim \mathcal{U}(\Omega)} \Big[ \| \mathbf{v}_t(\mathbf{x}) \|_2^2 \Big]$ penalizes the squared velocity of uniformly sampled points $\mathcal{U}(\Omega)$.

\section{Experiments and Results}
\subsection{Dataset and Implementation}

Our framework is  evaluated on two distinct datasets: 1) \textbf{SSM Dataset.}  Following~\cite{MIA23,MIA25}, we generated 10,000 3D left ventricular (LV)  meshes (9,000 for training, 1,000 for testing) using a 18-mode SSM. To compute the SDF ground truth, 100,000 spatial points were sampled per mesh.
2) \textbf{Control Cohort Dataset.}  We utilized a public  clinical Control Cohort~\cite{MIA25} comprising 144 normal transthoracic echocardiography subjects. Raw sequences were filtered for quality, temporally aligned to end-diastole, resized to $224\times224$, and the LV blood pool in apical views (A2C, A3C, A4C) was segmented via nnU-Net~\cite{nnunet}. 
\begin{table*}[t]
\centering
\caption{Quantitative comparison of  reconstruction performance, evaluating 3D geometric on SSM and 2D projection overlap on Control Cohort (CC) dataset. }
\label{tab:reconstruction_results}

\fontsize{8pt}{8pt}\selectfont  
\begin{tabular}{l cccc cc}
\toprule
\multirow{2}{*}{\textbf{Method}} & \multicolumn{4}{c}{\textbf{SSM
Dataset}} & \multicolumn{2}{c}{\textbf{CC Dataset}} \\
\cmidrule(lr){2-5} \cmidrule(lr){6-7}
& MAE(mm)$\downarrow$ & RMSE (mm)$\downarrow$ & HD (mm)$\downarrow$ & CD (mm)$\downarrow$ 
& Dice (\%)$\uparrow$ & IoU (\%)$\uparrow$ \\
\midrule

E-PiVox~\cite{Pix}
&3.41 ± 0.85 & 3.75 ± 1.05 & 13.04 ± 4.52 & 7.34 ± 1.88
& 87.35 ± 3.48 & 77.97 ± 4.02 \\

MIA`25~\cite{MIA25}     
& 2.61 ± 0.78 & 2.86 ± 0.86 & 7.98 ± 1.43 & 5.61 ± 0.89  
& 90.51 ± 2.72 & 84.76 ± 3.86 \\

\textbf{Echo4DIR}     
& 1.15 ± 0.13 & 1.33 ± 0.16 & 5.49 ± 0.54  & 2.84 ± 0.85 &  98.35 ± 0.42
& 96.75 ± 0.80  \\

\bottomrule
\end{tabular}
\label{tab1}
\end{table*}

\begin{figure}[t]
    \centering
    \begin{minipage}[c]{0.50\textwidth}
        \centering
        \captionof{table}{Quantitative Ablation. $M_x\rightarrow M_y$ denotes the reconstruction of  unseen view $M_y$ from input view $M_x$.}
        \label{tab:ablation_study}
      \fontsize{8pt}{8pt}\selectfont  
      \begin{tabular}{l c c}
    \toprule
    \multicolumn{3}{c}{\textbf{Effectiveness of ECA and DR}} \\
    \midrule
    Method & MAE (mm) $\downarrow$ & RMSE (mm) $\downarrow$  \\
    \midrule
    Echo4DIR                & 1.15 ± 0.13 & 1.33 ± 0.16  \\
    $w/o$ ECA                  & 1.28 ± 0.18   & 1.46 ± 0.21  \\
    $w/o$ DR                   & 1.54 ± 0.48
  & 1.74 ± 0.59  \\
    \midrule
    \midrule 
    
    \multicolumn{3}{c}{\textbf{Cross-View  Geometric Reliability}} \\
    \midrule
    Views (ch) & Dice (\%) $\uparrow$ & IoU (\%) $\uparrow$  \\
    \midrule
    4, 3 $\rightarrow$ 2 & 95.24 ± 1.40 & 90.94 ± 2.52 \\
    $w/o$ DR & 87.40 $\pm$ 5.37 & 81.87 $\pm$ 7.54 \\
    \midrule
    2 $\rightarrow$ 4, 3 & 94.89 ± 1.01 & 90.31 ± 1.81 \\
    $w/o$ DR & 84.73 $\pm$ 3.28  & 75.79 $\pm$ 4.33 \\
    \bottomrule
\end{tabular}

    \end{minipage}\hfill 
    \begin{minipage}[c]{0.50\textwidth}
        \centering
        \includegraphics[width=1.0\textwidth]{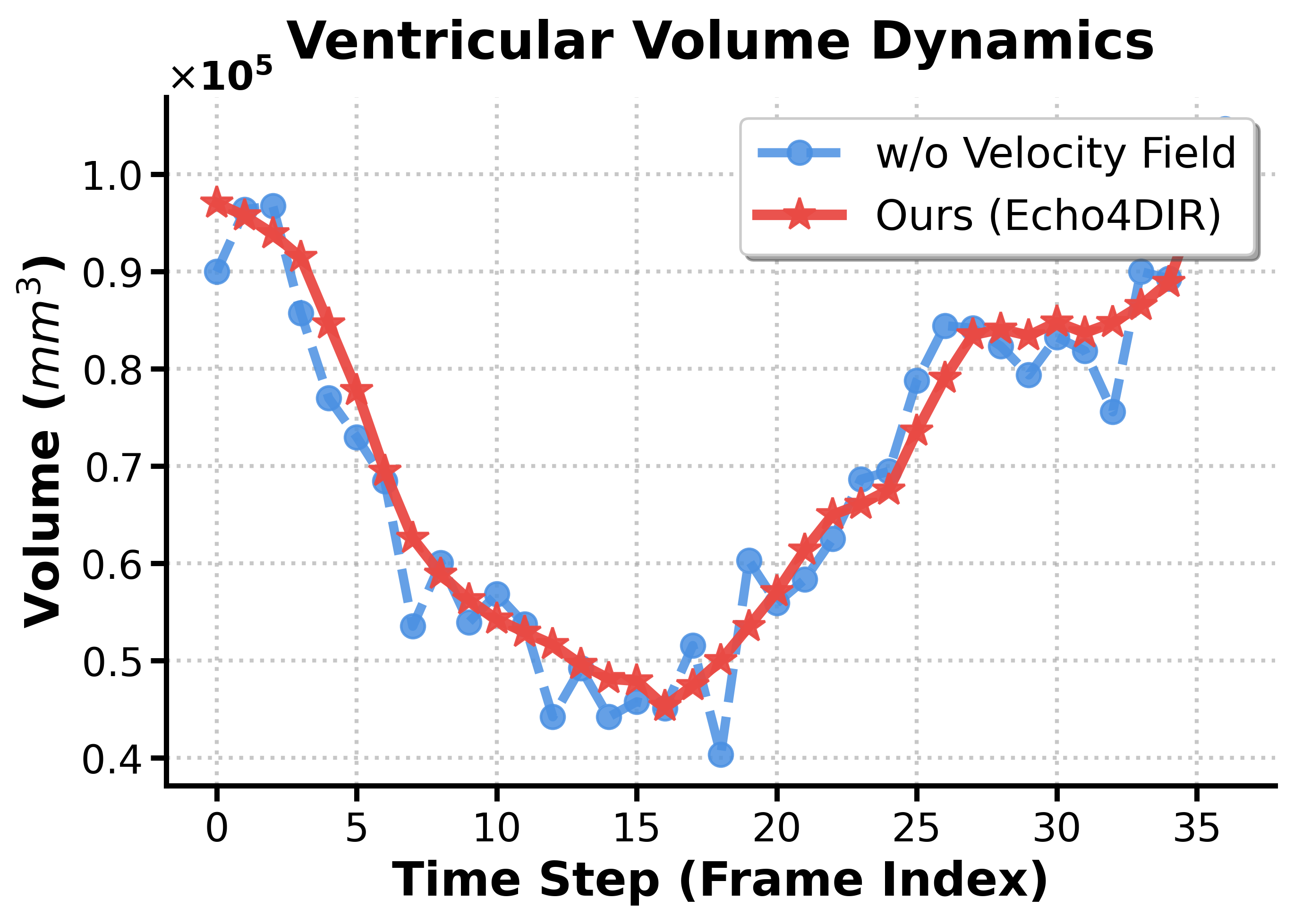}
        \caption{Comparison of  volume dynamics  with and without the velocity field.} 
        \label{fig2}
    \end{minipage}
\end{figure}

\textbf{Evaluation Metrics. }Our method is compared with the baseline approach MIA`25~\cite{MIA25}, the SOTA GCN-based method for LV mesh reconstruction. We evaluate geometric precision on SSM dataset using four standard 3D metrics: \textit{Mean Absolute Error} (MAE), \textit{Hausdorff Distance} (HD), \textit{Chamfer Distance} (CD), and \textit{Root Mean Squared Error} (RMSE), scaled to physical dimensions ($mm$).  For the CC dataset, we calculate the \textit{Dice Coefficient} (Dice) and \textit{Intersection over Union} (IoU) between the 2D projected masks and the clinical segmentation inputs.  

\textbf{Implementation Details. }Our framework is implemented in PyTorch on an RTX 4090 GPU. $\Phi_1$ and $\Phi_2$ employ ResNet-18 with dimensions of (64, 128) and (256, 512). The details of SDF follows~\cite{SDF}.  $\Psi_\omega$ employ a 4-layer MLP with positional encoding. During 3D+t TTO, we perform 1,500 steps to optimize the first frame as the anchor, using an alternating shape-probe optimization (8:2 time ratio) to prevent geometric distortion from poor initial poses. Probe parameters are frozen for the next 2,000 steps of shape-motion optimization (random frame selection).  We use the Adam optimizer with learning rates of $5\times10^{-3}$, $5\times10^{-4}$, and $1\times10^{-3}$ for the probe, latent shape, and velocity fields. $\lambda_{BCE}=1.0$, $\lambda_{Dice}=0.5$, $\lambda_{trans}=0.5$, and $\lambda_{smooth}=0.001$.  For rendering, 4,096 points are sampled per iteration, with the $\alpha$ increasing from 20 to 150.   3D and 4D reconstructions are completed within $\sim$ 30~s and 2~min.

\begin{figure}[t]
\includegraphics[width=\textwidth]{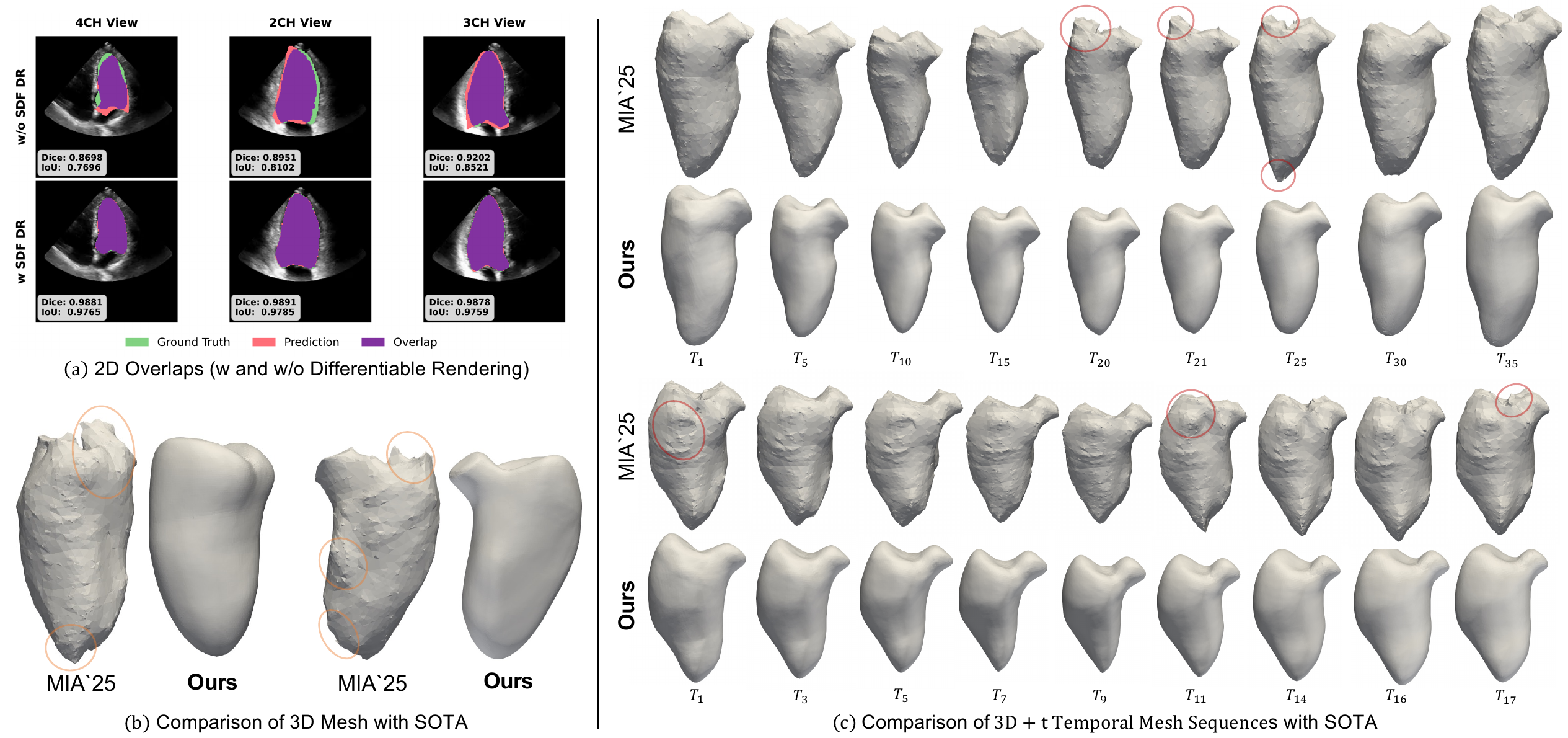}
\caption{Qualitative comparison of cardiac reconstruction performance.} 
\label{Fig3}
\end{figure}

\subsection{Quantitative and Qualitative Results}
Table~\ref{tab1} compares  Echo4DIR with state-of-the-arts, including voxel-based E-PiVox~\cite{Pix} and mesh-based MIA'25~\cite{MIA25}. On the SSM dataset, Echo4DIR significantly outperforms MIA'25 (MAE: $2.61$~$\rightarrow$$1.15$~mm). To evaluate the clinical utility, we further validated on the CC dataset. Echo4DIR achieves superior performance with a $98.35\%$ Dice. Such near-perfect overlap underscores the efficacy of our implicit representation in reconstructing geometric shape from clinical echo. In addition, Fig.~\ref{Fig3}~(b) and (c) visualize the 3D and 4D reconstructed meshes. Our method generates anatomically plausible cardiac surfaces with superior geometric smoothness, effectively eliminating unphysical artifacts and spikes. The 4D results capture consistent periodic cardiac motion.

\textbf{Ablation.} Table~\ref{tab:ablation_study} validates the necessity of Eipolar Cross-Attention (ECA) and Differentiable Rendering (DR). Omitting ECA increases RMSE from 1.33~mm to 1.46~mm, proving the necessity of cross-view correlations.  Removing DR causes RMSE to surge to 1.74~mm, demonstrating that gradient-based feedback is essential for robust shape adaptation.  Fig.~\ref{fig2} illustrates the intuitive overlap alignment, demonstrating DR`s effectiveness in processing real  clinical data. As shown in Fig.~\ref{Fig3}, the velocity field effectively smooths the volume curve, capturing the flat isovolumic relaxation  despite potential input segmentation errors.

\textbf{Geometric Reliability.} We further evaluate the reliability of our implicit prior by reconstructing unseen views from sparse inputs in Table~\ref{tab:ablation_study}. Notably, our framework achieves a remarkable 95.24\% Dice for these unseen views, demonstrating its superior reconstruction fidelity. In contrast, performance without DR collapses significantly, underscoring its critical role in ensuring cross-view consistency and reliable shape reconstruction.



\section{Conclusion}
This work presents Echo4DIR, a 4D implicit framework that bridges the gap between sparse 2D clinical observations and high-fidelity cardiac anatomical geometry. We integrate conditional SDFs with a test-time differentiable rendering strategy to achieve patient-specific shape adaptation, yielding a remarkable 98.35\% clinical Dice and 95.24\% Dice in unseen view reconstruction, which validates the superior geometric reliability.  The introduction of Radial SDF Alignment further ensures topological consistency during 3D+t extension, further eliminating non-physical artifacts.  Future work will extend this framework to whole-heart modeling and leverage large-scale CT or MRI priors for pretraining. In addition, we will  further explore the utility  and potential of these cardiac meshes for future digital twin integration such as surgical planning.

\begin{credits}
\subsubsection{\ackname} This work was supported by Singapore National Medical Research Council Open Fund - Young Individual Research Grant (25-1321-A0001), and Singapore Ministry of Education Tier 1 grant (25-1097-P0001). Y. Liu was partially supported by the China Scholarship Council (Grant No. 202506230034), supported by National Natural Science Foundation of China (Grant No.62162068, 62462066, 62466060), Yunnan Province Ten Thousand Talents Program and Yunling Scholars Special Project (Grant No. YNWR-YLXZ-2018-022), Joint Fund Project for "Double First-Class" Construction of Science and Technology Department of Yunnan Province and Yunnan University (Grant No. 202301BF070001-025), Scientific Research Fund of Yunnan Provincial Department of Education (No.2026Y0185, 2025J0008), supported by No. FWCY-BSPY2024009, No. KC-252513122.

\subsubsection{\discintname}
The authors have no competing interests to declare that are relevant to the content of this article.

\end{credits}


%
%
%
\bibliographystyle{splncs04}
\bibliography{mybibliography}

@inproceedings{yxh,
  title={4D myocardium reconstruction with decoupled motion and shape model},
  author={Yuan, Xiaohan and Liu, Cong and Wang, Yangang},
  booktitle={Proceedings of the IEEE/CVF International Conference on Computer Vision},
  pages={21252--21262},
  year={2023}
}

@inproceedings{Pix,
  title={Efficient Pix2Vox++ for 3D Cardiac Reconstruction from 2D echo views},
  author={Stojanovski, David and Hermida, Uxio and Muffoletto, Marica and Lamata, Pablo and Beqiri, Arian and Gomez, Alberto},
  booktitle={International Workshop on Advances in Simplifying Medical Ultrasound},
  pages={86--95},
  year={2022},
  organization={Springer}
}

@inproceedings{UltraTwin,
  title={UltraTwin: towards cardiac anatomical twin generation from multi-view 2D ultrasound},
  author={Yu, Junxuan and Duan, Yaofei and Huang, Yuhao and Wang, Yu and Ling, Rongbo and Luo, Weihao and Zhang, Ang and Xu, Jingxian and Ni, Qiongying and Zhou, Yongsong and others},
  booktitle={International Conference on Medical Image Computing and Computer-Assisted Intervention},
  pages={608--617},
  year={2025},
  organization={Springer}
}

@article{MIA23,
  title={Weakly supervised inference of personalized heart meshes based on echocardiography videos},
  author={Laumer, Fabian and Amrani, Mounir and Manduchi, Laura and Beuret, Ami and Rubi, Lena and Dubatovka, Alina and Matter, Christian M and Buhmann, Joachim M},
  journal={Medical image analysis},
  volume={83},
  pages={102653},
  year={2023},
  publisher={Elsevier}
}

@article{MIA25,
  title={2D echocardiography video to 3D heart shape reconstruction for clinical application},
  author={Laumer, Fabian and Rubi, Lena and Matter, Michael A and Buoso, Stefano and Fringeli, Gabriel and Mach, Fran{\c{c}}ois and Ruschitzka, Frank and Buhmann, Joachim M and Matter, Christian M},
  journal={Medical Image Analysis},
  volume={101},
  pages={103434},
  year={2025},
  publisher={Elsevier}
}

@inproceedings{DiT,
  title={Scalable diffusion models with transformers},
  author={Peebles, William and Xie, Saining},
  booktitle={Proceedings of the IEEE/CVF international conference on computer vision},
  pages={4195--4205},
  year={2023}
}

@inproceedings{STGCN,
  title={Spatial temporal graph convolutional networks for skeleton-based action recognition},
  author={Yan, Sijie and Xiong, Yuanjun and Lin, Dahua},
  booktitle={Proceedings of the AAAI conference on artificial intelligence},
  volume={32},
  number={1},
  year={2018}
}

@inproceedings{CycleGAN,
  title={Unpaired image-to-image translation using cycle-consistent adversarial networks},
  author={Zhu, Jun-Yan and Park, Taesung and Isola, Phillip and Efros, Alexei A},
  booktitle={Proceedings of the IEEE international conference on computer vision},
  pages={2223--2232},
  year={2017}
}

@inproceedings{SDF,
  title={Deepsdf: Learning continuous signed distance functions for shape representation},
  author={Park, Jeong Joon and Florence, Peter and Straub, Julian and Newcombe, Richard and Lovegrove, Steven},
  booktitle={Proceedings of the IEEE/CVF conference on computer vision and pattern recognition},
  pages={165--174},
  year={2019}
}

@inproceedings{ResNet,
  title={Deep residual learning for image recognition},
  author={He, Kaiming and Zhang, Xiangyu and Ren, Shaoqing and Sun, Jian},
  booktitle={Proceedings of the IEEE conference on computer vision and pattern recognition},
  pages={770--778},
  year={2016}
}

@book{MLP,
  title={Neural networks: a comprehensive foundation},
  author={Haykin, Simon},
  year={1994},
  publisher={Prentice hall PTR}
}

@inproceedings{ECA,
  title={ECSIC: Epipolar cross attention for stereo image compression},
  author={W{\"o}dlinger, Matthias and Kotera, Jan and Keglevic, Manuel and Xu, Jan and Sablatnig, Robert},
  booktitle={Proceedings of the IEEE/CVF Winter Conference on Applications of Computer Vision},
  pages={3436--3445},
  year={2024}
}

@article{NerF,
  title={Nerf: Representing scenes as neural radiance fields for view synthesis},
  author={Mildenhall, Ben and Srinivasan, Pratul P and Tancik, Matthew and Barron, Jonathan T and Ramamoorthi, Ravi and Ng, Ren},
  journal={Communications of the ACM},
  volume={65},
  number={1},
  pages={99--106},
  year={2021},
  publisher={ACM New York, NY, USA}
}

@incollection{Marching,
  title={Marching cubes: A high resolution 3D surface construction algorithm},
  author={Lorensen, William E and Cline, Harvey E},
  booktitle={Seminal graphics: pioneering efforts that shaped the field},
  pages={347--353},
  year={1998}
}

@inproceedings{BiLSTM,
  title={Hybrid speech recognition with deep bidirectional LSTM},
  author={Graves, Alex and Jaitly, Navdeep and Mohamed, Abdel-rahman},
  booktitle={2013 IEEE workshop on automatic speech recognition and understanding},
  pages={273--278},
  year={2013},
  organization={IEEE}
}

@inproceedings{La1,
  title={Occupancy flow: 4d reconstruction by learning particle dynamics},
  author={Niemeyer, Michael and Mescheder, Lars and Oechsle, Michael and Geiger, Andreas},
  booktitle={Proceedings of the IEEE/CVF international conference on computer vision},
  pages={5379--5389},
  year={2019}
}

@article{La2,
  title={Fronts propagating with curvature-dependent speed: Algorithms based on Hamilton-Jacobi formulations},
  author={Osher, Stanley and Sethian, James A},
  journal={Journal of computational physics},
  volume={79},
  number={1},
  pages={12--49},
  year={1988},
  publisher={Elsevier}
}

@article{nnunet,
  title={nnU-Net: a self-configuring method for deep learning-based biomedical image segmentation},
  author={Isensee, Fabian and Jaeger, Paul F and Kohl, Simon AA and Petersen, Jens and Maier-Hein, Klaus H},
  journal={Nature methods},
  volume={18},
  number={2},
  pages={203--211},
  year={2021},
  publisher={Nature Publishing Group US New York}
}

@ARTICLE{cai,
  author={Li, Lei and Camps, Julia and Jenny Wang, Zhinuo and Beetz, Marcel and Banerjee, Abhirup and Rodriguez, Blanca and Grau, Vicente},
  journal={IEEE Transactions on Medical Imaging}, 
  title={Toward Enabling Cardiac Digital Twins of Myocardial Infarction Using Deep Computational Models for Inverse Inference}, 
  year={2024},
  volume={43},
  number={7},
  pages={2466-2478},
  doi={10.1109/TMI.2024.3367409}}

@article{cf,
  title={Influence of excess fat on cardiac morphology and function: study in uncomplicated obesity},
  author={Iacobellis, Gianluca and Ribaudo, Maria Cristina and Leto, Gaetano and Zappaterreno, Alessandra and Vecci, Elio and Di Mario, Umberto and Leonetti, Frida},
  journal={Obesity research},
  volume={10},
  number={8},
  pages={767--773},
  year={2002},
  publisher={Wiley Online Library}
}

@article{cm,
  title={Cardiac morphology and blood pressure in the adult zebrafish},
  author={Hu, Norman and Yost, H Joseph and Clark, Edward B},
  journal={The Anatomical Record: An Official Publication of the American Association of Anatomists},
  volume={264},
  number={1},
  pages={1--12},
  year={2001},
  publisher={Wiley Online Library}
}

@article{CCT,
  title={Cardiac computed tomography},
  author={Boyd, Douglas P and Lipton, Martin J},
  journal={Proceedings of the IEEE},
  volume={71},
  number={3},
  pages={298--307},
  year={2005},
  publisher={IEEE}
}

@article{3DUS,
  title={A fast convolution-based methodology to simulate 2-Dd/3-D cardiac ultrasound images},
  author={Gao, Hang and Choi, Hon Fai and Claus, Piet and Boonen, Steven and Jaecques, Siegfried and Van Lenthe, G Harry and Van der Perre, Georges and Lauriks, Walter and D'hooge, Jan},
  journal={IEEE transactions on ultrasonics, ferroelectrics, and frequency control},
  volume={56},
  number={2},
  pages={404--409},
  year={2009},
  publisher={IEEE}
}

\end{document}